\documentclass[10pt,two column,letter paper]{article}

\usepackage{cvpr}              

\usepackage[dvipsnames]{xcolor}

\usepackage[accsupp]{axessibility}
\usepackage{courier}
\usepackage{graphicx}
\usepackage{array}
\usepackage{amsmath}
\usepackage{bbm}
\usepackage{booktabs}
\usepackage{graphicx}
\usepackage{float}
\usepackage{placeins}
\usepackage{bbding}
\usepackage{multirow}
\usepackage{booktabs}
\usepackage{makecell}
\usepackage{pifont}
\usepackage{stackengine}
\usepackage{caption}
\usepackage{overpic}
\usepackage{subcaption}
\definecolor{cvprblue}{rgb}{0.21,0.49,0.74}
\definecolor{lightblue}{rgb}{0.91,0.94,0.98}
\usepackage[pagebackref,breaklinks,colorlinks,citecolor=cvprblue]{hyperref}

\usepackage[capitalize]{cleveref}
\crefname{section}{Sec.}{Secs.}
\Crefname{section}{Section}{Sections}
\Crefname{table}{Table}{Tables}
\crefname{table}{Tab.}{Tabs.}

\def\paperID{13898} 
\def\confName{CVPR}
\def\confYear{2024}

\newcommand{\tabref}[1]{Tab. \ref{#1}}

\newcommand{\figref}[1]{Fig. \ref{#1}}

\newcommand{\myPara}[1]{\vspace{5pt}\noindent$\bullet$~\textbf{#1} \quad}

\def\eg{\emph{e.g.}}

\def\etal{{\em et al.~}}

\title{LAKE-RED: Camouflaged Images Generation by Latent Background Knowledge Retrieval-Augmented Diffusion}

\author{
~~~Pancheng Zhao$^{1,2}$~~~~~~~~~Peng Xu$^3$\thanks{Corresponding Author.}~~~~~~~~~~Pengda Qin$^4$~~~~~~~~Deng-Ping Fan$^{2,1}$\\
Zhicheng Zhang$^{1,2}$~~~~~~~~Guoli Jia$^1$~~~~~~~~~Bowen Zhou$^3$~~~~~~~~~Jufeng Yang$^{1,2}$\\
$^1$ VCIP \& TMCC \& DISSec, College of Computer Science, Nankai University\\
$^2$ Nankai International Advanced Research Institute (SHENZHEN· FUTIAN)\\
$^3$ Department of Electronic Engineering, Tsinghua University ~~~~~~
$^4$ Alibaba Group\\
\hspace{-8pt}{\tt\small{pc.zhao99@gmail.com, peng\_xu@tsinghua.edu.cn, pengda.qpd@alibaba-inc.com, dengpfan@gmail.com}}\\
\hspace{-6pt}{\tt\small{gloryzzc6@sina.com, exped1230@gmail.com, zhoubowen@tsinghua.edu.cn, yangjufeng@nankai.edu.cn}}
}

\begin{document}
\maketitle
\begin{abstract}
\vspace{-5pt}
Camouflaged vision perception is an important vision task with numerous practical applications.
Due to the expensive collection and labeling costs, this community struggles with a major bottleneck that the species category of its datasets is limited to a small number of object species.
However, the existing camouflaged generation methods require specifying the background manually, thus failing to extend the camouflaged sample diversity in a low-cost manner. 
In this paper, we propose a Latent Background Knowledge Retrieval-Augmented Diffusion (LAKE-RED) for camouflaged image generation.
To our knowledge, our contributions mainly include:
(1) For the first time, we propose a camouflaged generation paradigm that does not need to receive any background inputs.
(2) Our LAKE-RED is the first knowledge retrieval-augmented method with interpretability for camouflaged generation, in which we propose an idea that knowledge retrieval and reasoning enhancement are separated explicitly, to alleviate the task-specific challenges.
Moreover, our method is not restricted to specific foreground targets or backgrounds, offering a potential for extending camouflaged vision perception to more diverse domains.
(3) Experimental results demonstrate that our method outperforms the existing approaches, generating more realistic camouflage images. 
%
\textbf{\textcolor{black}{Our source code is released on \href{https://github.com/PanchengZhao/LAKE-RED}{https://github.com/PanchengZhao/LAKE-RED}.}}


\end{abstract}

\section{Introduction}
\vspace{-5pt}
\vspace{-5pt}
\myPara{Background.}
Camouflaged vision perception \cite{fan2023advances} 
is a challenging problem (\eg, camouflaged object detection~\cite{fan2021concealed}) aiming to perceive the concealed complex patterns and extensively applied in various fields such as pest detection \cite{ebrahimi2017vision}, healthcare \cite{yuan2023full,huang2024annotationefficient,ji2024rethinking}, and autonomous driving \cite{burnett2019autotrack, liu2024active, wang2024multi, AlignSAM, jiaming2024learning}.
%
%
It has made significant progress in recent years.
However, these kinds of overly complex visual scenes and patterns make it extremely time-consuming and labor-intensive to annotate the pixel-wise masks.
There is a fact that an instance-level annotation in the COD10K dataset took an average of 60 minutes \cite{fan2020camouflaged}, far longer than the 3 minutes in the COCO-Stuff dataset \cite{caesar2018coco}, clearly illustrating this issue.
Thus, this community struggles with a major bottleneck in that the species category of its datasets is limited to a small number of object species, \eg, animals. 

%
%
%
%

\begin{figure}[t!]
    \centering
    \includegraphics[width=.98\linewidth]{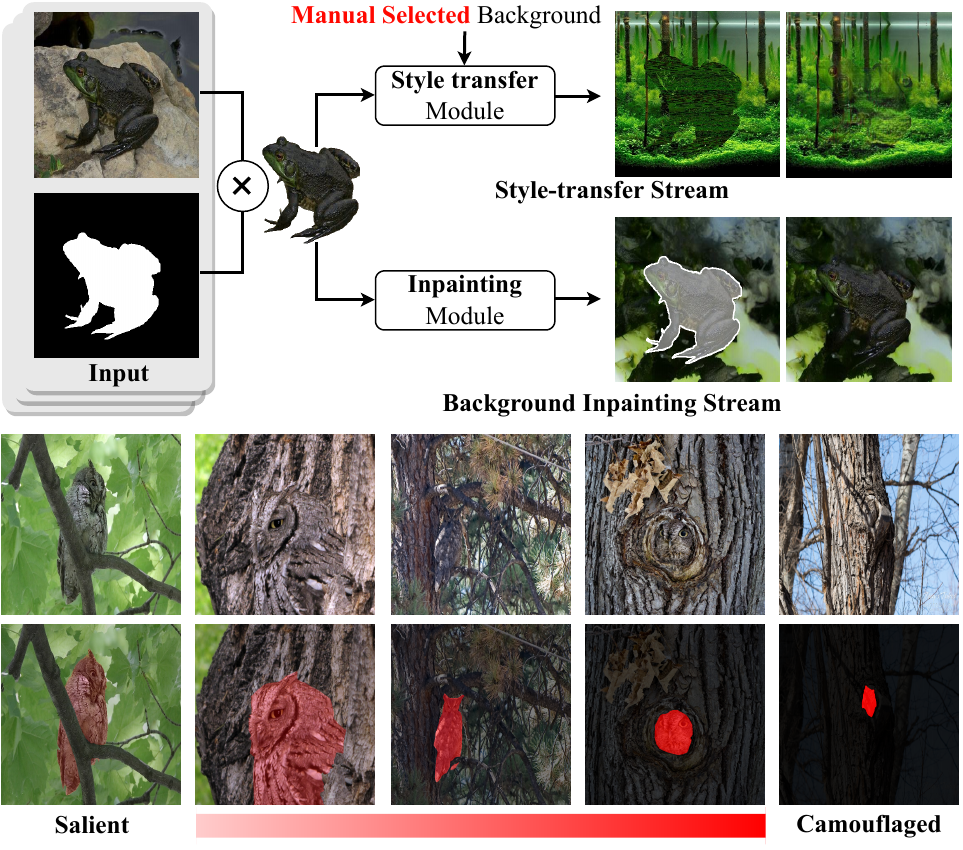}
    \vspace{-5pt}
    \caption{\small \textbf{Comparison of Frameworks for Camouflage Image Generation.}
    %
    Existing methods rely on manually specified backgrounds, which not only receive limitations in diversity and scope from the human's own cognition but also result in expensive image generation on a large scale.
    %
    Without changing the texture of itself, the same target can be camouflaged to different degrees in different environments.
    Inspired by it, we synthesize camouflaged images through a background inpainting stream, hiding by automatically choosing a suitable background for the object.
    } 
    \vspace{-10pt}
    \label{fig:motivation}
\end{figure}

%
\vspace{-5pt}
\myPara{Existing Technical Limitations.}
Recently, the rapid development in the AIGC community, particularly generative models based on GAN \cite{creswell2018generative} and Diffusion \cite{ho2020denoising}, has revealed the potential of using synthetic data to address data scarcity.
%
%
DatasetGAN \cite{zhang2021datasetgan} and BigDatasetGAN \cite{li2022bigdatasetgan} train a shallow decoder to generate pixel-level annotations from the feature space of pre-trained GANs. 
%
DiffuMask \cite{wu2023diffumask} is inspired by the attention map in the Diffusion Model and obtains pixel-level annotations from the cross-attention process of the text and image.
However, the above method is designed for generic scenarios, and the generated data has a significant domain gap with the training data for the camouflage vision perception task.
Moreover, as shown in \figref{fig:motivation}, the existing camouflaged generation methods require specifying the background manually, thus failing to extend the camouflaged sample diversity in a low-cost manner. 
%
%

\vspace{-5pt}
\myPara{Motivation.}
Our idea is to make full use of the domain-specific traits of camouflaged scenes to implement a low-cost solution.
As shown in \figref{fig:motivation}, the level of target camouflage depends largely on its surrounding environmental context.
 Furthermore, we observed that a majority of camouflaged images utilize a background-matching perceptual deception strategy, where the concealed object blends seamlessly into the surrounding background. 
In this scenario, the foreground and background regions of the camouflaged image exhibit remarkable visual perceptual consistency. 
For instance, the frog concealed in the grass surface displays a mottled pattern of green and brown just like the grass and ground. 
This feature convergence between foreground and background makes it possible to retrieve and reason about background through foreground features.
%

\vspace{-5pt}
\myPara{Method Overview.}
Inspired by the above motivation, we introduce LAKE-RED, a pipeline that automatically generates high-quality camouflage images and pixel-level segmentation masks.
%
%
The model accepts a foreground object input to achieve object-to-background image inpainting. 
Specifically, the model first perceives features from the foreground and utilizes them as queries to \textit{retrieve latent background knowledge} from a pre-constructed external knowledge base.
%
%
\textcolor{black}{Then, the model learns to \textit{reason} from foreground objects to background scenes via using the retrieved knowledge to conduct the camouflaged background reconstruction. 
This helps the model achieve a richer condition-guided background generation.}
%
%
Simultaneously, this synthesis preserves the precise foreground annotation and prevents boundary blurring caused by mask generation.
\figref{fig:gallery} illustrates pairs of camouflaged images generated by our LAKE-RED, along with examples of two application scenarios. Without the need for manually specified background inputs, the proposed model can efficiently produce high-quality camouflaged images at a low cost.

\vspace{-5pt}
\myPara{Contribution.}
(1) For the first time, we propose a camouflaged generation paradigm without any background inputs.
(2) Our LAKE-RED is the first knowledge retrieval-augmented method with interpretability for camouflaged generation, in which we propose an idea that knowledge retrieval and reasoning enhancement are separated explicitly, to alleviate the task-specific challenges.
Moreover, our method is not restricted to specific foregrounds or backgrounds, offering a potential for extending camouflaged vision perception to more diverse domains.
(3) Experimental results demonstrate our method outperforms the existing approaches, generating more realistic camouflage images.

\section{Related Work}
\vspace{-5pt}
%

%
\myPara{Synthetic Dataset Generation.}
Synthetic data has gained significant attention as one of the primary approaches to tackle data bottlenecks in deep learning methods due to its low cost \cite{kar2019meta,mumuni2024survey}. 
Previous research on synthetic datasets has mainly focused on producing high-quality simulated scenes in 3D environments and generating data from them, which has been extensively employed for tasks such as recognition \cite{zhao2021emotion,wang2022ease,jia2022s,wen2023dip}, segmentation \cite{wrenninge2018synscapes,10097456,chen2023confidence,liu2023meaningful}, object tracking \cite{Zhang_2023_ICCV,liu2023progressive}, image and video understanding \cite{9472932,Zhang_2023_CVPR,10.1145/3503161.3548007,zhang2024masked,zhang2024distribution,zhai2024looking}, optical flow estimation \cite{butler2012naturalistic,liu2023pgfnet}, and 3D reconstruction \cite{zhou2020towards,zhou2021dc,zhou2024AST}.
The considerable disparity between the distribution of synthetic data through simulated scenarios and real data restricts their validity.
Significant progress in generative modeling has recently enabled the reduction of the domain gap between synthetic and real data. 
With realistic image data generated by advanced generative models (\eg, GAN, DALL-E2, and Stable Diffusion), some research has attempted to investigate the potential of synthetic data as a replacement for real data \cite{he2022synthetic,ge2022dall,li2023guiding}.
Specifically, DatasetGAN \cite{zhang2021datasetgan} and BigDatasetGAN \cite{li2022bigdatasetgan} excel in generating a significant quantity of synthetic images with segmentation masks with limited labeled data. 
On the other hand, Diffumask \cite{wu2023diffumask} relies exclusively on textual supervision to extract semantic labels from the cross-attention maps of text and images.

\myPara{Camouflage Image Generation.}
Camouflage images are different from regular images as they contain one or more concealed objects \cite{fan2020camouflaged}. 
Although the concept of camouflage can be traced back to Darwin's theory of evolution \cite{stevens2009animalcam,cuthill2019camouflage,merilaita2017howcamwork} and has long been used in various fields, the task of camouflage image generation was not proposed until 2010 by Chu~\etal \cite{chu2010camouflage}. 
The proposed model gets a specified foreground and background as input and uses hand-crafted features to give the foreground textural details similar to the background, making the concealed objects difficult for humans to recognize. 
Recent advancements in deep learning methods for style transfer and image composing have provided new ideas for generating camouflage images.
Subsequent models, such as Zhang's \cite{zhang2020deep} and Li's \cite{li2022location}, have further improved camouflage image generation by composing the foreground with the background through style transfer and structure alignment.
However, the use of artificially specified backgrounds increases the cost of data acquisition and limits the diversity of generated images due to human cognitive limitations.
These limitations make it impossible to generate large-scale datasets, greatly reducing the application value of the generated images.
\begin{figure*}[t!]
    \vspace{-22pt}
    \centering
    \includegraphics[width=\linewidth]{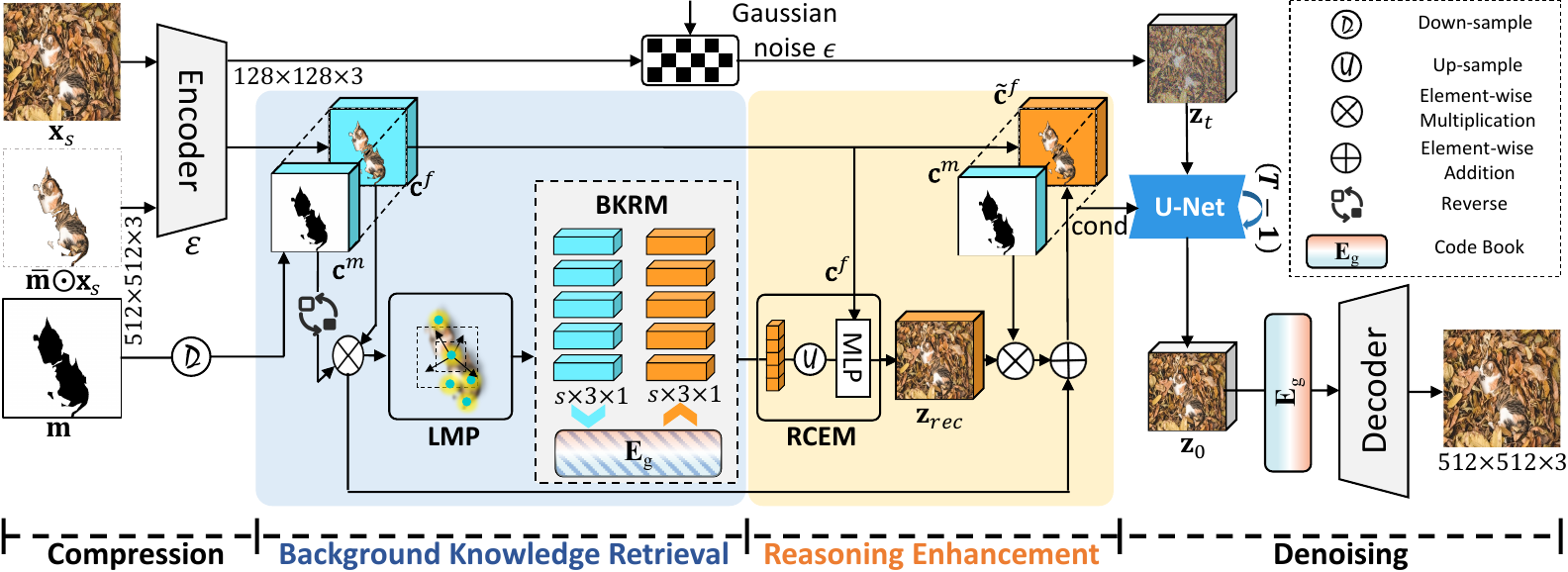}
    \vspace{-15pt}
    \caption{\small \textbf{The pipeline of our camouflaged images generation framework LAKE-RED.}
    Our framework mainly includes three steps: 
    (1) Extracting visual representations of foreground areas by Localized Masked Pooling (LMP).
    (2) The Background Knowledge Retrieval Module (BKRM) is utilized to retrieve background-related features from the codebook.
    %
    (3) The Reasoning-Driven Condition Enhancement module (RCEM) allows the model to learn foreground-to-background reasoning through a background reconstruction.
    %
    %
    }
    \vspace{-10pt}
    \label{fig:pipeline}
\end{figure*}
%
%
\section{Methodology}
\vspace{-5pt}
Our objective is to generate camouflaged images by automatically complementing the background region for a specific foreground object, resulting in a realistic image where the object is concealed in the generated background.
While there have been advancements in camouflage image generation methods, manually specifying the background is not practical due to the high human cost and limited cognitive range.
Through our observation of the camouflage phenomenon, we have noticed that the background region of a camouflaged image often shares similar image features with the surface of the foreground object. This suggests that a suitable camouflage background may already exist within the foreground image itself.
%
%
Formally, given a source image $ \mathbf{x}_s\in \mathbb{R}^{H\times W\times 3} $, containing an object with an irregular shape. 
The object's location is precisely indicated by a binary mask $\mathbf{m}$ with the same size as the original image $\mathbf{x}_s$, where $\mathbf{m}_{i,j}=0$, with $i \in [0, H]$ and $j \in [0, W]$, represents the object region that needs to be maintained in subsequent operations, and $\mathbf{m}_{i,j}=1$ represents the editable background region.
The model takes $\left \{ \mathbf{x}_s, \mathbf{m}  \right \} $ as input, and outputs a camouflaged image $\mathbf{x}_c$.
The objective is to obtain a prior from the foreground $\mathbf{x}_s \odot  \mathbf{\Bar{m}}$ to generate a suitable background that replaces the original one. The foreground should harmoniously match the new background.

%
\subsection{Preliminaries}
\vspace{-5pt}
\myPara{Revisiting Latent Diffusion Models.}
Aiming to generate high-quality camouflage images, our proposed method is based on classic Latent Diffusion Models (LDM) ~\cite{rombach2022high}.
Similar to other probabilistic models, LDM learns the probability distribution $p(x)$ of a given image set $x$ through self-supervised training and achieves high-quality image generation by reversing a Markov forward process.
Specifically, the forward process adds a sequence noise to the original images $\mathbf{y}_0=\mathbf{x_s}$ to obtain a noisy image $\left \{ \mathbf{y}_t\mid t\in \left [ 1, T \right ]     \right \} $, where $\mathbf{y}_t=\alpha_t\mathbf{y}_0+(1-\alpha_t)\mathbf{\epsilon} $.
As $\alpha_t$ decreases with time step $t$, more Gaussian noise $\epsilon$ is introduced into $\mathbf{y}_0$.
The generation process can be described as a sequence of denoising autoencoders $\epsilon_\theta \left ( \mathbf{y}_t, \mathbf{c}, t  \right ) $ to predict a denoised variant of input $\mathbf{y}_t$.
Furthermore, in order to decrease the computational demands of high-resolution image synthesis for the model, a pre-trained autoencoder $\varepsilon $ is employed to encode $\mathbf{y}$ into a latent representation $\mathbf{z} = \varepsilon \left ( \mathbf{y}  \right ) $, where $\mathbf{z} \in \mathbb{R}^{h\times w\times c}$.
So the training objective can be defined as the following loss function:
\begin{equation}
\setlength{\abovedisplayskip}{3pt}
\setlength{\belowdisplayskip}{3pt}
\mathcal{L}=\mathbb{E}_{t,\varepsilon \left ( \mathbf{y}  \right ) ,\mathbf{\epsilon}}\left \| \epsilon_\theta \left ( \mathbf{z}_t, \mathbf{c}, t  \right )- \mathbf{\epsilon}  \right \|  _{2}^{2}.
\end{equation}
For the inpainting stream, the condition $\mathbf{c}$ includes $\mathbf{x}_s\odot \mathbf{\Bar{m}}$ to indicate the remaining area. 
Once $T$ steps have been completed, the model predicts the latent representation $\mathbf{z}^{\prime}_0$, of which the noise $\epsilon$ has been entirely removed.
Finally, to reconstruct a high-resolution image from the latent representation, a VQVAE \cite{van2017neural} based decoder $\mathcal{D}$ is utilized in the final stage.
The visual information from the code book $\mathbf{e}$ is embedded into the latent representation by incorporating a quantization layer $\nu$ into the decoder, which can be yielded as:
\begin{equation}
\setlength{\abovedisplayskip}{3pt}
\setlength{\belowdisplayskip}{3pt}
\mathbf{y}^{\prime}_0= \mathcal{D} \left ( \nu (\mathbf{e},\mathbf{z}^{\prime}_0 ) \right ) ,
\end{equation}
where $\mathbf{e} \in \mathbb{R}^{K\times D}$, $K$, and $D$ denote the size of the discrete latent space and the dimensionality of each latent embedding vector, respectively.

\subsection{Model Designs}
\vspace{-5pt}
Current image inpainting methods accept a conditional input $\mathbf{c}$ that includes known image regions and indicates editable regions, which can be defined as:
\begin{equation}
\setlength{\abovedisplayskip}{3pt}
\setlength{\belowdisplayskip}{3pt}
\begin{aligned}
\mathbf{c}^f, \mathbf{c}^m &= \varepsilon (\mathbf{I}_{known} ),downsample({\mathbf{m} }, f),\\
\mathbf{c} &= Concat\left ( \mathbf{c}^f, \mathbf{c}^m \right ),
\end{aligned}
\end{equation}
where $\mathbf{I}_{known} = \mathbf{x}_s \odot  \mathbf{\Bar{m}}$, and $\mathbf{m}$ is down sampled by a factor $f=2^n$, with $n \in \mathbb{N} $.
However, they tend to prioritize preserving the structural continuity of the object in the image and infer to fill in the missing areas.
The inference of the model is constrained when the non-edited region forms a complete object that lacks structural continuity with the background. 
This means that the current condition is not enough to facilitate the model in making accurate inferences from the foreground object to the background scene. 
%
To mitigate the negative impact of this performance bottleneck on the results, as shown in \figref{fig:pipeline}, we focus on retrieving richer background knowledge and develop a reasoning-based background reconstruction task that enables the model to explicitly learn the relationship between the foreground and background of a camouflaged image. 
The reconstructed features can then be used to enhance existing conditions and provide the model with richer guidance information.
%
\subsubsection{Background Knowledge Retrieval}
\vspace{-5pt}
As mentioned before, inferring from object to background is a significant challenge for image inpainting models. 
However, unlike general images, camouflage images are primarily characterized by background matching, where the background and the object exhibit a high degree of consistency in terms of texture. 
%
This implies that it becomes feasible to retrieve background knowledge using foreground features.
The training framework for reconstructing backgrounds through masked ground truth (GT) implicitly models the relationship between the object and background, which results in the model paying insufficient attention to the texture consistency of the object and background. 
%
Explicitly retrieving background features aligned with the object features is a viable option to provide richer guidance for the denoising process.
In order to obtain feature representations about the background texture, we take inspiration from the autoencoder and decoder used by LDM, which is based on VQ-VAE.
VQ-VAE constructs a code book $\mathbf{e}$ in the embedding space between the encoder and the decoder during the training process. 
The codebook can be injected with features into the representation of the latent space by vector quantized operation before the decoder to obtain a better reconstruction performance.
To address the issue of missing background features of the condition, the pre-trained codebook is replicated and shifted to the denoising process as a global visual embedding $\mathbf{E}_{g} = \mathbf{e}^{\mathrm{T}} \in \mathbb{R} ^{D\times K} $. The process of obtaining background features $\mathbf{x}^b$ using a latent space codebook $E_{g}$ can be summarized as:
\begin{equation}
\setlength{\abovedisplayskip}{3pt}
\setlength{\belowdisplayskip}{3pt}
\begin{aligned}
    \mathbf{x}^{b} &= \text{Concat}(\mathbf{h}_1, \mathbf{h}_2, \dots, \mathbf{h}_H)\mathbf{W}^{f\to b}, \\
    \mathbf{h}_i &= a_i\mathbf{E}_g\mathbf{W}_{i}^{V}, \\ 
    a_i &= \text{softmax}\left(\frac{\left[\mathbf{x}^f\mathbf{W}_{i}^{Q}\right] \cdot \left[\mathbf{E}_g\mathbf{W}_{i}^{K}\right]^\mathrm{T}}{\sqrt{d_k}}\right).
\end{aligned}    
\end{equation}
We feed the foreground feature $\mathbf{x}^f$ into the Multi-Head Attention (MHA) layer with $H$ heads, as the query, for retrieving the related background content from codebook $\mathbf{E}_{g}$, and obtain the background aligned visual feature $\mathbf{x}^b$.

\subsubsection{Localized Masked Pooling}
\vspace{-5pt}
We introduce a simple and efficient latent background knowledge retrieval module, denoted as $\mathcal{B}  \left ( \mathbf{x}^f, \mathbf{E}_g \right ) $, that retrieves background-aligned visual features $\mathbf{x}^b$ from codebook $\mathbf{E}_g$ using foreground features $\mathbf{x}^f$.
The richness of the feature representation $\mathbf{x}^f$ extracted from $\mathbf{c}^f$ directly impacts the validity of features that can be retrieved from the codebook.
Thus, the foreground feature representation $\mathbf{x}_f$ can become another potential performance bottleneck.
To exclude features in the background region during feature extraction, a straightforward approach is to follow \cite{zhang2020sg} using Masked Averaged Pooling (MAP), to obtain representative vectors of foreground features as:
\begin{equation}
\setlength{\abovedisplayskip}{3pt}
\setlength{\belowdisplayskip}{3pt}
\mathbf{x}^f_i = \Phi \left ( \mathbf{c}^f_i, \mathbf{c}^m \right ) = \frac{ {\textstyle \sum_{x=1,y=1}^{w,h}}\mathbf{c}^f_{i,x,y}\ast \bar{\mathbf{c}}^m_{x,y} }{ {\textstyle \sum_{x=1,y=1}^{w,h}{\bar{\mathbf{c}}^m_{x,y}} } }  ,
\end{equation}
where $i \in \{1,2,\dots, \vartheta \}$ indicates the channel number.
The MAP treats the foreground as a whole and compresses it into a unified representation, which can lead to a significant loss of information. 
In particular, the encoder $\varepsilon \left ( \cdot  \right ) $ maintains the channel number of the feature to be $3$, resulting in $\mathbf{x}^f \in \mathbb{R}^{3\times 1} $. 
%
This simple representation is insufficient to capture the rich features of the foreground and can limit the effectiveness of latent background knowledge retrieval.
Foreground objects in camouflaged images often display intricate visual features, which we define as a combination of $s$ sub-features. The higher the value of $s$, the more intricate and detailed the corresponding feature is.
To extract richer foreground features, we shift our focus from global to local and employ the SLIC algorithm \cite{achanta2012slic} to cluster the foreground regions into $s$ superpixels.
The above process can be reformulated as:
\begin{equation}
\setlength{\abovedisplayskip}{3pt}
\setlength{\belowdisplayskip}{3pt}
\begin{aligned}
    \mathbf{p}^i_1,\mathbf{p}^i_2,\cdots ,\mathbf{p}^i_s & = \mathcal{S} (\mathbf{c}^f_i, \mathbf{c}^m),\\
    \mathbf{x}^f_{i,j} & = \Phi_s \left ( \mathbf{p}^i_j, \mathbf{c}^m \right ) = \frac{ {\textstyle \sum_{x = 1,y = 1}^{w,h}\mathbf{c}^f_{i,x,y}}\ast \mathbf{p}^i_{j,x,y} }{ {\textstyle \sum_{x = 1,y = 1}^{w,h} {\mathbf{p}^i_{j,x,y}} } }.
\end{aligned}
\end{equation}
\subsubsection{Reasoning-Driven Condition Enhancement}
\vspace{-5pt}
Additionally, we upsample the obtained background knowledge features $\mathbf{x}^b$ and combine them with the foreground features $\mathbf{c}^f$ to reconstruct the GT image features $\mathbf{z}_0 = \varepsilon (\mathbf{y} _0), \mathbf{z}_0 \in \mathbb{R} ^{h\times w\times c}$.
The reconstruction feature can be computed as:
\begin{equation}
\setlength{\abovedisplayskip}{3pt}
\setlength{\belowdisplayskip}{3pt}
\begin{aligned}
    \mathbf{z}_{rec}= \mathbf{MLP}  ( Concat(\mathbf{c}^f,upsample(\mathbf{x}^b, f))).
\end{aligned}
\end{equation}
Then, $\mathbf{z}_{rec}$ is utilized to refine the initial condition of the input. 
To emphasize the background features, we created a feature reconstruction task that enhances the model's ability to reason about real background features using background knowledge.
Specifically, we populate the background region of $\mathbf{c}^f$ with the reconstructed $\mathbf{z}_{rec}$ to strengthen the information embedded in the condition while reserving the foreground areas.
The strategy for enhancing the condition can be formulated as:
\begin{equation}
\setlength{\abovedisplayskip}{3pt}
\setlength{\belowdisplayskip}{3pt}
\begin{aligned}
    \tilde{\mathbf{c}}^f & = \mathbf{c}^f \cdot (1 - \mathbf{c}^m) + \mathbf{z}_{rec} \cdot \mathbf{c}^m,\\
    \mathbf{\tilde{c}} &= Concat\left ( \Tilde{\mathbf{c}}^f, \mathbf{c}^m \right ).
\end{aligned}
\end{equation}
For the loss of background reconstruction, we have:
\begin{equation}
\setlength{\abovedisplayskip}{3pt}
\setlength{\belowdisplayskip}{3pt}
    \mathcal{L}_{bgrec} = \frac{1}{h \times w} \sum_{i=1}^{h} \sum_{j=1}^{w} (\mathbf{z}_{rec} \cdot \mathbf{c}^m   - \mathbf{z}_{0}\cdot \mathbf{c}^m)^2.
\end{equation}
Then, the overall loss can be reformulated as:
\begin{equation}
\setlength{\abovedisplayskip}{3pt}
\setlength{\belowdisplayskip}{3pt}
\begin{aligned}
    \mathcal{L} &= \mathcal{L}_{diff} + \mathcal{L}_{bgrec}\\
    &\propto \left \| \epsilon_\theta \left ( \mathbf{z}_t, \mathbf{\tilde{c} }, t  \right )- \epsilon  \right \|  _{2}^{2}+\left \| \mathbf{z}_{rec} \cdot \mathbf{c}^m   - \mathbf{z}_{0}\cdot \mathbf{c}^m \right \|^2 .
\end{aligned}
\end{equation}
By leveraging the properties of the camouflaged image, we refine and enhance the input condition $\mathbf{c}$. 
While defining the image features of the foreground area, the enhanced condition $\mathbf{\tilde{c}}$ guides the generation of background. 
The implicit and explicit constraints work together to help the model learn the texture consistency between the foreground object and the background, resulting in high-quality camouflage image generation.

\begin{figure*}[t!]
    \vspace{-20pt}
    \centering
    \begin{overpic}
        [width=\linewidth]{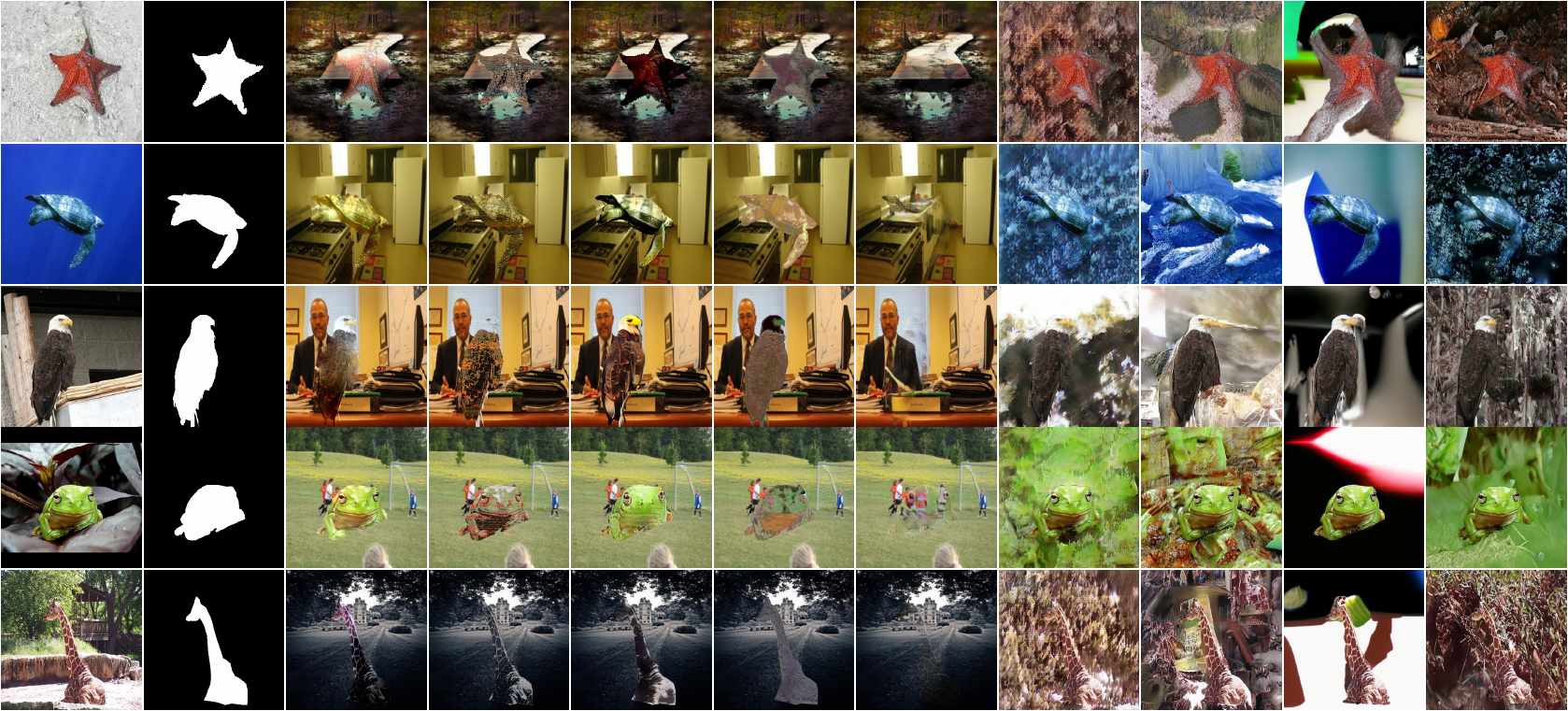}
        \put(94,-1.8){{\small Ours}}
        \put(82,-1.8){{\small Repaint \cite{lugmayr2022repaint}}}
        \put(73.6,-1.8){{\small LDM \cite{rombach2022high}}}
        \put(64.8,-1.8){{\small TFill \cite{zheng2022bridging}}}
        \put(54.5,-1.8){{\small LCGNet \cite{li2022location}}}
        \put(46.8,-1.8){{\small DCI \cite{zhang2020deep}}}
        \put(36.8,-1.8){{\small AdaIN \cite{huang2017arbitrary}}}
        \put(29.8,-1.8){{\small CI \cite{chu2010camouflage}}}
        \put(19.8,-1.8){{\small AB \cite{perez2003poisson}}}
        \put(11.8,-1.8){{\small Mask}}
        \put(2.3,-1.8){{\small Image}}
    \end{overpic}
    \vspace{-5pt}
    \caption{\small \textbf{Comparison with existing methods in transferring general images into camouflaged images.}
    The first two columns are the input images and we provide camouflaged images generated by nine methods for the comparison.
    Note that the methods in columns 3 to 7 additionally share a randomly sampled background image as input.
    }
    \vspace{-10pt}
    \label{fig:compare}
\end{figure*}

\section{Experiments}
\vspace{-5pt}
\subsection{Experimental Setups}
\vspace{-5pt}
\myPara{Datasets.}
%
%
Following the previous works \cite{fan2021concealed} for COD, 4,040 images (3,040 from COD10K \cite{fan2020camouflaged}, 1,000 from CAMO \cite{le2019anabranch}) are used as real data for training the model.
To verify the generative performance, we collected image-mask pairs from various fields to construct a test data set, including three subsets: Camouflaged Objects (CO), Salient Objects (SO), and General Objects (GO).
In CO, there are 6,473 pairs of images from CAMO \cite{le2019anabranch}, COD10K \cite{fan2020camouflaged}, and NC4K \cite{lv2021simultaneously}.
%
Then we randomly selected 6473 images from the well-known salient object detection datasets (DUTS \cite{wang2017learning}, DUT-OMRON \cite{yang2013saliency},  etc.) and the segmentation dataset (COCO2017 \cite{lin2014microsoft}) to evaluate the performance of the model on open domain data.
%

%
\myPara{Metrics.}
Following the good practices of AIGC \cite{li2022bigdatasetgan,rombach2022high} and COD \cite{luo2023camdiff,lamdouar2023making}, we choose the InceptionNet-based metrics FID \cite{binkowski2018demystifying} and KID \cite{heusel2017gans} to measure the quality of generated camouflaged images.
%
%
Once the image features are extracted by InceptionNet, the distance between them is computed to indicate the level of resemblance between the model's output and the target dataset.
Different from the general images, well-synthesized camouflaged images should not include easily identifiable objects, and it is more challenging to extract discriminative features \cite{luo2023camdiff}.
%
%
%
A smaller value indicates that the generated image is more similar to the real camouflaged image.
\myPara{Implementation Details.}
To generate camouflaged images by given foreground images, we utilize a powerful Latent Diffusion Model \cite{rombach2022high} pre-trained in the inpainting task as initialization. 
The model is designed to handle images and masks of size $512\times512$ and is compressed to a latent space of $128\times128\times3$ using a pre-trained VQVAE~\cite{van2017neural}.
During training, we focus on training the denoising U-Net and do not fine-tune the auto-encoder and decoder components. 
We refine and enhance the existing condition through the proposed module in this paper. 
The parameters optimization such as initialization, data augmentation, and batch size are set similar to the original paper.
Finally, the model generates the camouflaged image and resizes it to align with the original input.
We conduct all the experiments by PyTorch and GeForce RTX 3090 GPUs are used for all experiments.
\subsection{Comparison with the State-of-the-art Methods}
\vspace{-5pt}
Previous camouflage image generation methods are based on image blending or style transfer, which differ fundamentally from the method proposed in this paper. 
Thus, for each solution, we select cutting-edge methods for comparison.
For the image blending and style transfer schemes, the model requires a manually specified background image when accepting a foreground input. 
We used Places365~\cite{zhou2017places}, a large-scale scene dataset, as the source of background images. 
For a given foreground input, we randomly sampled a background image from Places365, resized it, and then performed image synthesis process. 
To facilitate comparison between different methods, all methods shared the same background image for a given foreground input.
For the image inpainting scheme, the model only accepts one foreground input and generates a camouflaged image as output.
\begin{table*}[t!]
    \vspace{-15pt}
    \footnotesize
    \centering
     \renewcommand{\arraystretch}{1.1}
     \setlength\tabcolsep{1pt}
    \caption{\small \textbf{Quantitative performance.} 
    The proposed camouflaged image generation method is subjected to a quantitative evaluation, wherein it is compared with state-of-the-art (SOTA) methods. The evaluation involved specific foreground objects sampled from camouflaged images, salient images, and general images. The proposed method shows excellent performance.
    } 
    \vspace{-5pt}
    \begin{tabular}{p{3cm}<{\centering}p{2.4cm}>{\raggedleft\arraybackslash}p{2.3cm}<{\centering}p{1.1cm}<{\centering}p{1.1cm}<{\centering}p{1.1cm}<{\centering}p{1.1cm}<{\centering}p{1.1cm}<{\centering}p{1.1cm}<{\centering}p{1.1cm}<{\centering}p{1.1cm}<{\centering}p{1.1cm}<{\centering}}
    \toprule[1.2pt]
    \multicolumn{2}{c}{\multirow{2}{*}{Methods}} &\multirow{2}{*}{Input} & \multicolumn{2}{c}{Camouflaged Objects}&\multicolumn{2}{c}{Salient Objects}&\multicolumn{2}{c}{General Objects}&\multicolumn{2}{c}{Overall}\\
    \cmidrule(lr){4-5} \cmidrule(lr){6-7} \cmidrule(lr){8-9} \cmidrule(lr){10-11} 
    && &FID$\downarrow$&KID$\downarrow$&FID$\downarrow$&KID$\downarrow$&FID$\downarrow$&KID$\downarrow$&FID$\downarrow$&KID$\downarrow$ \\
    \hline
    \specialrule{0em}{1pt}{0pt}
    \multirow{5}{*}{\shortstack{\textbf{\textit{{Image}}}\\\textit{\textbf{Blending}}}}& AB ~\cite{perez2003poisson}$_{03}$& $\mathcal{F}+\mathcal{B}$ & 117.11 & 0.0645 & 126.78 & 0.0614 & 133.89 & 0.0645 & 120.21 & 0.0623 \\
    & CI \cite{chu2010camouflage}$_{10}$& $\mathcal{F}+\mathcal{B}$ & 124.49 & 0.0662 & 136.30 & 0.7380 & 137.19 & 0.0713 &  128.51 & 0.0693 \\
    & AdaIN ~\cite{huang2017arbitrary}$_{17}$& $\mathcal{F}+\mathcal{B}$ & 125.16 & 0.0721 & 133.20 & 0.0702 & 136.93 & 0.0714 & 126.94 & 0.0703  \\
    & DCI \cite{zhang2020deep}$_{20}$& $\mathcal{F}+\mathcal{B}$ & 130.21 & 0.0689 & 134.92 & 0.0665 & 137.99 & 0.0690 & 130.52 & 0.0673 \\
    & LCGNet \cite{li2022location}$_{22}$& $\mathcal{F}+\mathcal{B}$ & 129.80 & 0.0504 & 136.24 & 0.0597 & 132.64 & \textbf{0.0548} & 129.88 & 0.0550 \\
    \hline
    \multirow{4}{*}{\shortstack{\textit{\textbf{Image}}\\\textit{\textbf{Inpainting}}}}& TFill ~\cite{zheng2022bridging}$_{22}$& $ \mathcal{F}$ & 63.74 & 0.0336 & 96.91 & 0.0453 & 122.44 & 0.0747 & 80.39 & 0.0438  \\
    & LDM ~\cite{rombach2022high}$_{22}$& $ \mathcal{F}$ & 58.65 & 0.0380 & 107.38 & 0.0524 & 129.04 & 0.0748 & 84.48 & 0.0488  \\
    & RePaint-L~\cite{lugmayr2022repaint}$_{22}$& $ \mathcal{F}$ & 76.80 & 0.0459 & 114.96 & 0.0497 & 136.18 & 0.0686 & 96.14 & 0.0498 \\
    & Ours$_{23}$ & $ \mathcal{F}$ & \textbf{39.55} & \textbf{0.0212} & \textbf{88.70} & \textbf{0.0428} & \textbf{102.67} & 0.0625 & \textbf{64.27} & \textbf{0.0355} \\
    \bottomrule[1pt]
    \end{tabular}
    \vspace{-10pt}
    \label{tab:method_compared}
\end{table*}
%

%
\myPara{Qualitative analysis.}
\figref{fig:compare} presents a comparison of the quality of camouflaged images generated by our method and other methods from a general image.
The results show that methods such as AB and CI are highly influenced by the background image input, despite the foreground features being processed to align with the background. 
As a result, the foreground exhibits conflicts with the background scenes and objects, such as the eagle and turtle in the room (2nd and 3rd rows), and the larger-than-life frog (4th row).
LCGNet performs the best in hiding the objects, with their features being barely visible. 
Camouflaged objects in nature are seamlessly embedded in the background rather than being completely invisible.
On the other hand, image inpainting methods only require foreground object input and adaptive background generation can meet the requirements of large-scale generation. 
However, existing methods suffer from issues such as lack of authenticity of the background (TFill), low degree of camouflage (LDM), and failure of background complementation (Repaint-L).
In contrast, our method naturally integrates the given target into the generated background, preserving all the target's features while achieving overall camouflage of the image.
\myPara{Quantitative analysis.}
A large-scale test set is constructed to evaluate the quality of camouflage image generation, which includes three types of foreground objects to assess the model's adaptability to different image domains.
The salient objects subset and the general objects subset are sampled from datasets in the salient object detection and image segmentation domains, respectively, with the number of images kept consistent with the COD test set.
The distance between the generated results and the real COD benchmarks is measured using FID and KID, and the results are presented in \tabref{tab:method_compared}.

The results on the three subsets display a step-wise distribution, indicating that the model performance was strongly influenced by the image domain gap, with general objects being more challenging to transform than salient objects.
The image blending-based methods produce large results because they mechanically shift the foreground features towards being consistent with the background features, resulting in image visual features that are primarily determined by the background image. 
When the background image is randomly sampled, the related indexes also exhibit some degree of randomness.
On the other hand, image inpainting-based schemes tended to generate a suitable background for the object and generally show better performance.

In addition, we observe outliers in the validation results of LCGNet on the subset of General Objects, which are caused by a combination of the following reasons.
First, the difficulty of synthesizing increases in three subsets. 
The camouflaged object comes from a concealed scene and is easy to hide. 
The salient object is of moderate size and position and usually has a complete structure.
The general object has a rich variety of classes and diverse sizes, making it challenging to find suitable camouflage environments for it.
As the complexity rises, these approaches progressively struggle to conceal general objects flawlessly, leading to a decline in performance within that particular subset.
In this case, LCGNet maximally discards foreground features, and the results mainly depend on the randomly sampled backgrounds (\figref{fig:compare}). 
It is least affected by the negative influence from the foreground and is introduced to randomness by the background, thus resulting in anomalous results.
%
However, our method achieved optimal performance on the overall test set.
\begin{figure*}[t!]
    \vspace{5pt}
    \centering
    \includegraphics[width=\linewidth]{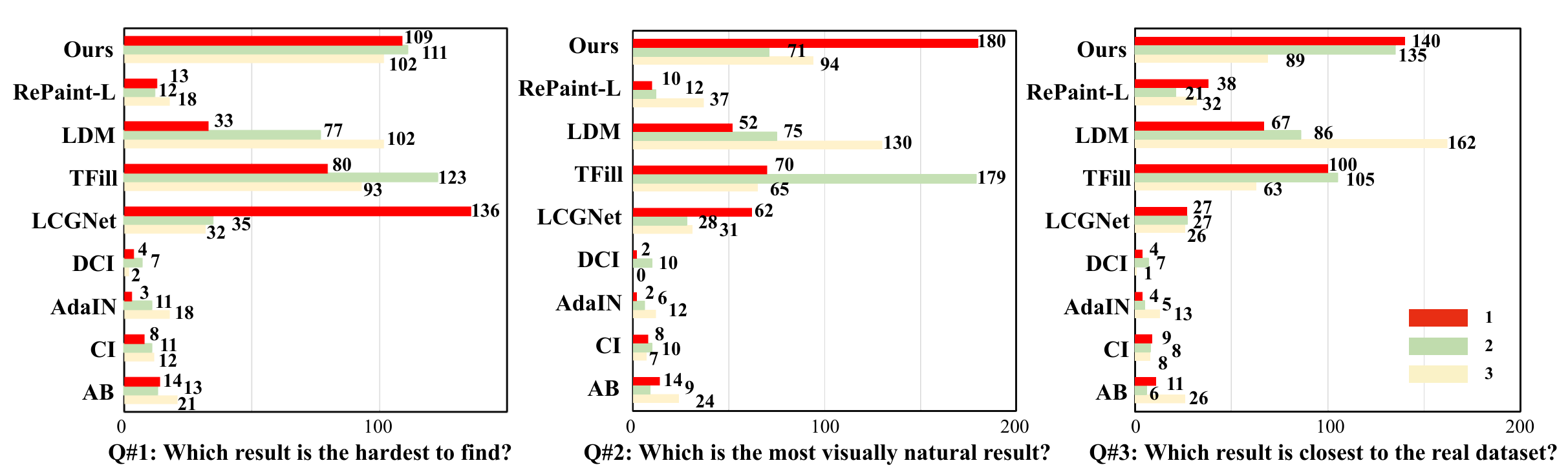}
    \caption{\small \textbf{User study about subjective ratings of the camouflaged image generated by 9 different methods.}
    Our method is considered to produce the most natural and visually closest results to the real camouflage image.
    }
    \label{fig:userstudy}
\end{figure*}

\begin{figure*}[t!]
    \vspace{5pt}
    \centering
    \begin{overpic}
        [width=\linewidth]{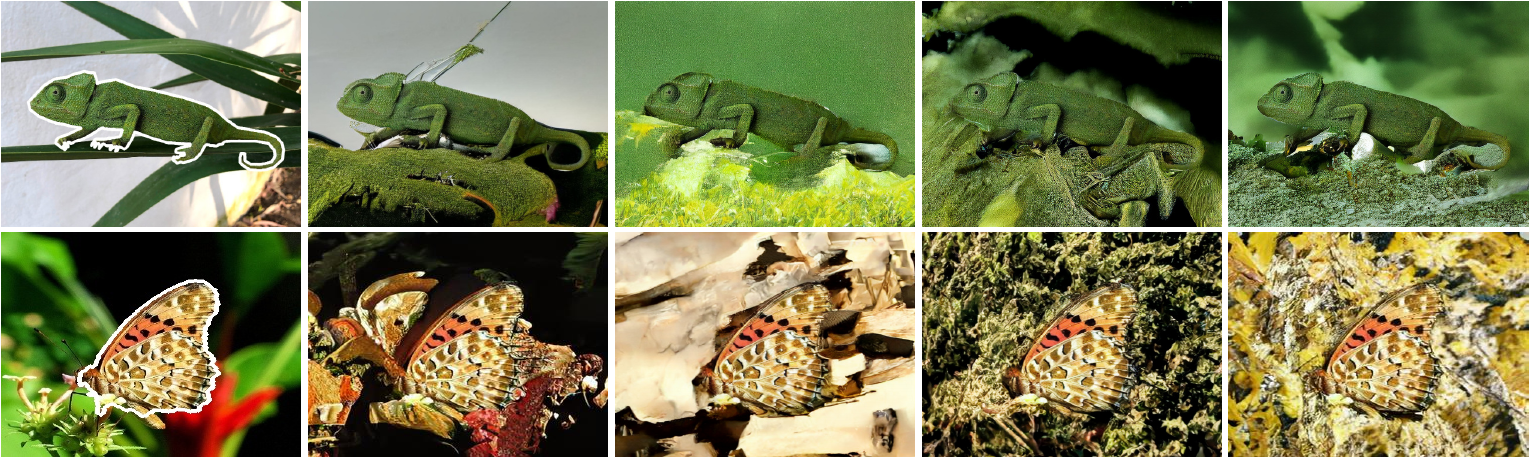}
        \put(3.5, -1.5){{\small Foreground Image}}
        \put(28,-1.5){{\small Base}}
        \put(47,-1.5){{\small +BKRM}}
        \put(63.5,-1.5){{\small +BKRM+RCEM}}
        \put(81,-1.5){{\small +BKRM+RCEM+LMP}}
    \end{overpic}
    \caption{\small \textbf{The visualization of ablation study.}
    We visualize the samples during the ablation experiments to visualize the effectiveness of the modules we proposed.
    }
    \vspace{-10pt}
    \label{fig:ablation}
\end{figure*}
\myPara{User Study.}
Since both image generation quality and camouflage effectiveness require human perception, we conducted user studies to obtain subjective human judgments on the generated results. 
To this end, we followed the previous work on camouflage image generation to randomly select 20 sets of foreground images and applied various methods to generate the results. 
For style transfer-based methods, we used an additional image randomly sampled from Places365 as the background input, which was kept consistent for all methods.
We invited 20 participants to rate the results based on three questions:

\noindent~\textbf{-Q\#1: Which result is the hardest to find? }

\noindent~\textbf{-Q\#2: Which is the most visually natural result?}

\noindent~\textbf{-Q\#3: Which result appears closest to the real camouflaged image dataset?}

For each question, participants need to select their top 3 choices, with 1 being the highest.
The results of the user survey are presented in \figref{fig:userstudy}. Although LCGNet received more votes in \textbf{Q\#1} due to the almost invisible foreground in the generated results, our method was considered to produce more natural and visually closer results to the real dataset in terms of visual presentation.
\begin{table}[h!]
    \scriptsize
    \centering
     \renewcommand{\arraystretch}{1.1}
     \setlength\tabcolsep{1.6pt}
    \caption{\small \textbf{Quantitative Ablation study.} 
    We progressively add each module to the base model to compare their impact on the quality of the generated results and costs.
    The result shows that the method we proposed is effective and almost cost-free.
    } 
    \vspace{-5pt}
    \begin{tabular}{
    |p{.7cm}<{\centering}
    p{.7cm}<{\centering}
    p{.7cm}<{\centering}|
    p{1.1cm}<{\centering}
    p{1cm}<{\centering}
    p{1cm}<{\centering}|
    p{1cm}<{\centering}
    p{1cm}<{\centering}|
    }
        \hline
        \multicolumn{3}{|c|}{\textbf{Module}} & \multirow{2}{*}[+0ex]{\textbf{Prams(M)$\downarrow$}} & \multirow{2}{*}[+0ex]{\textbf{MAC(G)$\downarrow$}} & \multirow{2}{*}[+0ex]{\textbf{FPS(Hz)$\uparrow$}} & \multicolumn{2}{c|}{\textbf{Overall}} \\ \cline{1-3}\cline{7-8}
        BKRM & RCEM & LMP &  &  &  & FID$\downarrow$ & KID$\downarrow$\\
        \hline
        \ding{55} & \ding{55} & \ding{55} & 440.46 & 577.97 & 0.2482 & 96.14 & 0.0498   \\
        \checkmark & \ding{55} & \ding{55} & 440.47 & 577.99 & 0.2442 & 69.80 & 0.0417   \\
        \checkmark & \checkmark & \ding{55} & 440.47 & 577.99 & 0.2438 & 69.52 & 0.0412   \\
        \checkmark & \checkmark & \checkmark &  440.47  & 577.99 & 0.2008 & 64.27 & 0.0355  \\
        \hline
    \end{tabular}
    \vspace{-5pt}
    \label{tab:ablation}
\end{table}
\subsection{Ablation Study}
\vspace{-5pt}
We conduct the ablation study by gradually adding modules to the base LDM to evaluate the effectiveness of each component in our proposed method. 
As shown in \tabref{tab:ablation}, the quality of the generated camouflage images gradually improves with the introduction of the modules proposed in this paper, demonstrating the effectiveness.
When all three modules are applied simultaneously, the model performance reaches its peak, achieving improvements of 33.14\% and 28.71\% in the FID and KID metrics, respectively.
At this point, the introduction of the three modules only adds about 0.01M parameters and 0.02G of computation to the model, with the inference speed reduced by only 0.04Hz.
These results clearly indicate that our method is effective and comes at almost no additional cost.

%
We further visualize the samples during the ablation experiments to show the effectiveness of these modules.
Two sets of results are shown in \figref{fig:ablation}.
The LDM faces challenges in focusing on the camouflage properties during inpainting from the foreground object to the background. 
It also struggles to generate the background in certain regions, resulting in black color blocks due to the complexity of the task.
By incorporating a latent background knowledge retrieval module (BKRM), the model is explicitly constrained to learn foreground and background similarity, resulting in a closer alignment of the generated background with the foreground.
%
Furthermore, the reasoning-driven condition enhancement module (RCEM) enhances the realism of scenes by incorporating a background reconstruction loss that compels the model to reason and reconstruct the background features accurately.
Finally, the introduction of localized masked pooling (LMP) shifted the model's attention from global to local foreground features, enhancing the texture diversity of the generated background.
\section{Conclusion}
\vspace{-5pt}
We propose a latent background knowledge retrieval-augmented diffusion (LAKE-RED) for camouflaged image generation.
Unlike existing methods, our generation paradigm is background-free.
By knowledge retrieval and reasoning enhancement, we get a strong background condition from the foreground, resulting in synthetic images that surpass those generated by other SOTA camouflaged image generation methods.
Our approach is not restricted to specific foreground targets or human-selected backgrounds. 
This enables us to generate camouflage images on a large scale and offers the potential for extending camouflaged vision perception to more diverse domains in the future.
%

{\small
\bibliographystyle{ieeenat_fullname}
\bibliography{egbib}
}
\end{document}